\documentclass[sigconf]{acmart-me}

\usepackage{booktabs} 
\usepackage{url}
\usepackage{color}
\usepackage{enumitem}
\usepackage{subcaption}
\usepackage{balance}
\usepackage{multirow}

\setcopyright{rightsretained}

\acmDOI{}

\acmISBN{}

\acmConference[MediaEval'21]{Multimedia Evaluation Workshop}{December 12-15 2021}{Online} 
\acmYear{2021}
\copyrightyear{2021}

\acmPrice{}

\begin{document}
\title{Deep Models for Visual Sentiment Analysis of Disaster-related Multimedia Content}

\author{Khubaib Ahmad\textsuperscript{1}, Muhammad Asif Ayub\textsuperscript{1}, Kashif Ahmad\textsuperscript{2},\\ Ala Al-Fuqaha\textsuperscript{2}, Nasir Ahmad\textsuperscript{1}}
\affiliation{\textsuperscript{1} Department of Computer Systems Engineering, University of Engineering and Technology, Peshawar, Pakistan. \\ \textsuperscript{2} Division of Information and Computing Technology, College of Science and Engineering, Hamad Bin Khalifa University, Qatar Foundation, Doha, Qatar.}
\email{{khubaibtakkar,asifayub836}@gmail.com, {kahmad,aalfuqaha}@hbku.edu.qa, n.ahmad@uetpeshawar.edu.pk}  

%
%
%
%
%

\renewcommand{\shortauthors}{K. Ahmad et al.}
\renewcommand{\shorttitle}{Visual Sentiment Analysis: A Natural Disaster Use-case}

\begin{abstract}
This paper presents a solutions for the MediaEval 2021 task namely ''Visual Sentiment Analysis: A Natural Disaster Use-case''. The task aims to extract and classify sentiments perceived by viewers and the emotional message conveyed by natural disaster-related images shared on social media. The task is composed of three sub-tasks including, one single label multi-class image classification task, and, two multi-label multi-class image classification tasks, with different sets of labels. In our proposed solutions, we rely mainly on two different state-of-the-art models namely, Inception-v3 and VggNet-19, pre-trained on ImageNet, which are fine-tuned for each of the three task using different strategies. Overall encouraging results are obtained on all the three tasks. On the single-label classification task (i.e. Task 1), we obtained the weighted average F1-scores of 0.540 and 0.526 for the Inception-v3 and VggNet-19 based solutions, respectively. On the multi-label classification i.e., Task 2 and Task 3, the weighted F1-score of our Inception-v3 based solutions was 0.572 and 0.516, respectively. Similarly, the weighted F1-score of our VggNet-19 based solution on Task 2 and Task 3 was 0.584 and 0.495, respectively.

\end{abstract}

%
%
%
%
%


\maketitle

\section{Introduction}
\label{sec:intro}
Over the last few years, natural disasters analysis in social media outlets has been one of the active areas of research. During this time several interesting solutions exploring different aspects of natural disasters have been proposed \cite{said2019natural}. Some key aspects of natural disasters explored in the literature include disaster detection \cite{said2021active}, disaster news dissemination \cite{ahmad2017jord}, and disasters severity and damage assessment \cite{alam2020deep,kankanamge2020determining}. Some efforts on the  sentiment analysis of natural disaster-related social media posts have also been reported. However, most of the efforts made in this regard are based on textual information \cite{beigi2016overview}. More recently, Hassan et al. \cite{hassan2019sentiment} introduced the concept of visual sentiment analysis of natural disaster-related images by proposing a deep sentiment analyzer. However, the topic is very challenging and there are several aspects of visual sentiment analysis of natural disaster-related visual content that yet need to be explored. As part of their efforts to further explore the topic, the authors proposed a task namely ''Visual Sentiment Analysis:  A Natural Disaster Use-case Task at MediaEval 2021'' \cite{hassan2021vissentiment}. 

This paper provides the details of the solutions proposed by team CSE-Innoverts for the visual sentiment analysis task. The task is composed of three sub-tasks including a (i) single-label multi-class classification task with three labels, a (ii) multi-label multi-class classification task with seven labels, and a (iii) multi-label multi-class classification task with 11 labels. In the first task, the participants need to classify an image into \textit{Negative}, \textit{Positive}, and \textit{Neutral} sentiments. In the second task, the proposed solution aims to differentiate among \textit{Joy}, \textit{sadness}, \textit{fear}, \textit{disgust}, \textit{anger}, \textit{surprise}, and \textit{neutral}. The final task is composed of 11 labels including \textit{anger}, \textit{anxiety}, \textit{craving}, \textit{empathetic pain}, \textit{fear}, \textit{horror}, \textit{joy}, \textit{relief}, \textit{sadness}, and \textit{surprise}.  

\section{Proposed Approach}
\subsection{Methodology for Single-label Classification task (Task 1)}
For the first task, we mainly rely on two different Convolutional Neural Networks (CNNs) architectures namely Inception V-3 and VggNet based on their proven performances in similar tasks \cite{said2018deep}. Since the available dataset is not large enough to train the models from the scratch, we fine-tuned the existing models pre-trained on ImageNet dataset \cite{deng2009imagenet}. In the literature, generally, the models pre-trained on ImageNet and Places dataset \cite{zhou2017places} are fine-tuned for image classification tasks. However, our choice for the current implementation is based on the better performance of the models pre-trained on the ImageNet dataset in similar tasks \cite{said2018deep}. In this work, the models are fine-tuned for 50 epochs using Adam optimizer with a learning rate of 0.0001.  

It is important to mention that the provided dataset is imbalanced with a large number of \textit{negative} samples while fewer samples are available in the \textit{neutral} class. Before fine-tuning the models, we applied an up-sampling technique to balance the dataset. Moreover, some data augmentation techniques are also employed to further increase the training samples by cropping, rotating, and flipping the image patches.  

\subsection{Methodology for Multi-label Classification tasks (Task 2 and 3)}
We used the same strategy of fine-tuning existing pre-trained state-of-the-art models for Task 2 and Task 3. However, to deal with the multi-label classification, several changes are made. For instance, the top layers of the models are extended to support the multi-label classification tasks. Moreover, the sigmoid Cross-Entropy loss function is used to deal with every CNN output vector component independently. Moreover, the up-sampling technique is used to balance the dataset. The data augmentation techniques are also employed in these tasks. 

\section{Results and Analysis}
\subsection{Evaluation Metric}
The evaluations of the proposed solutions are carried out in terms of weighted F1-score, which is also the official metric of evaluation of the task.

\subsection{Experimental Results on the development set}

Table \ref{Dev_results} provides the experimental results of our proposed solutions on the development set in terms of F1-score. As can be seen, overall better results are obtained on the single-label classification Task 1, which is composed of three classes only. As we go deeper in the sentiment categories/classes hierarchy the performance of the algorithms decreases as the inter-class variation decreases.  As far as the performance of the models is concerned, Inception-v3 has significant improvements over VggNet-19 on Task 1 and Task 2 while comparable results are obtained on Task 3.

\begin{table}[!ht]
\caption{Evaluation of our proposed solutions on the development set in terms of F1-score.}
\label{Dev_results}
\begin{tabular}{|l|l|l|l|}
\hline
\textbf{Runs} & \textbf{Task 1} & \textbf{Task 2} & \textbf{Task 3} \\ \hline
 Run 1 &  0.714 & 0.588 & 0.479  \\ \hline
Run 2& 0.666& 0.535& 0.479  \\ \hline

\end{tabular}
\end{table}


\subsection{Experimental Results on the test set}

Table \ref{Test_results} presents the official results of our proposed solutions on the test set. Surprisingly, overall better results are obtained on a multi-label classification task Task 2 for both the models. On the other hand, similar to the development set, the least performance is observed for both models on Task 3. As far as the performance of the models is concerned, Inception-v3 based solution outperformed the VggNet-19 based solution on Task 1 and Task 3 while comparable results are obtained on Task 2.
\begin{table}[!ht]
\caption{Evaluation of our proposed solutions on the test set in terms of weighted F1-score.}
\label{Test_results}
\begin{tabular}{|l|l|l|l|l|}
\hline
\textbf{Runs} & \textbf{Task 1} & \textbf{Task 2} & \textbf{Task 3} \\ \hline
Run 1 &  0.540 & 0.572 & 0.516 \\ \hline
Run 2& 0.526 & 0.584 & 0.495 \\ \hline
\end{tabular}
\end{table}

\section{Conclusions and Future Work}
The challenge is composed of three tasks including a single-label and two multi-label image classification tasks with different sets of labels. The first task aims to cover the conventional three categories/labels generally used to represent sentiments. The other two tasks aim to cover sets of labels more specific to natural disasters. These three sets of labels allow to explore different aspects of the domain, and the task's complexity increases by going deeper in the sentiments hierarchy. For all the tasks, we rely on two state-of-the-art deep architectures namely Inception-v3 and VggNet-19. To this aim, the models pre-trained on the ImageNet dataset are fine-tuned on the development dataset. 

In the current implementations, we rely on object-level information only by employing the models pre-trained on ImageNet dataset. We believe, scene-level features could also contribute to the task. In the future, we aim to jointly utilize both object and scene-level information for better performance on all the tasks. Moreover, we aim to employ merit-based fusion schemes by considering the contribution of the individual models to the tasks. 
\balance

\bibliographystyle{ACM-Reference-Format}
\def\bibfont{\small} 
\bibliography{sigproc} 

\end{document}